\documentclass[a4paper,fleqn]{cas-dc}

\usepackage[numbers]{natbib}
\usepackage{booktabs}

\begin{document}
\let\WriteBookmarks\relax
\def\floatpagepagefraction{1}
\def\textpagefraction{.001}

\shorttitle{Structural Negative Transfer in Federated GNNs}
\shortauthors{C. P. Kabgere, S. SS}

\title [mode = title]{Structural Negative Transfer in Federated Graph Neural Networks: Diagnosis, Causal Investigation, and the Limits of Divergence-Aware Mitigation}

\author[1,2]{Chethana Prasad Kabgere, Shylaja SS}
\cormark[1]
\ead{}

\affiliation[1]{organization={PES University},
    city={Bengaluru},
    country={India}}

\affiliation[2]{organization={PES University},
    city={Bengaluru},
    country={India}}

\cortext[cor1]{Corresponding author}

\begin{abstract}
Federated learning lets multiple participants train one shared model without pooling their raw data, by exchanging locally trained model updates instead of the data itself. The standard method for combining those updates, federated averaging, assumes that averaging locally trained models is a reasonable way to solve one shared problem, provided the participants' data are broadly similar. A decade of work on non-IID federated learning has shown this assumption survives real strain, including differing label and feature distributions, without breaking. We ask whether it survives a different kind of strain specific to graph neural networks, in which client graphs differ not in label or feature distribution but in structure itself, so that the same shared weights must operate over fundamentally different topologies. We call the resulting harm structural negative transfer. In a federation of real citation networks and synthetic structural proxies, a structurally atypical client lost more than half its achievable accuracy simply by joining. In an initial six client federation, two label free structural statistics computable before training were strongly associated with this harm. Expanding the federation to twenty clients produced a more precise result. Degree divergence remained associated with harm, weaker but still significant, and it survived the removal of all domain contrast. Spectral divergence did not replicate, and we trace that failure to a confound in how leave one out statistics depend on the composition of the reference pool. We report this alongside an extended attempt to explain and fix the surviving association, weighting failure and success equally. A causal intervention isolating topology finds no significant effect. A degree normalization mechanism holds across twenty four seeds but does not explain the harm when corrected. The best of five candidate fixes beats a tuned baseline only until a matched, structurally blind control is applied, at which point the gain disappears. What survives is a modest, partially replicated, degree specific signal that is not yet a validated predictor at scale.
\end{abstract}


\maketitle

\section{Introduction}

Machine learning models generally improve with more data, and the most straightforward way to get more data is to pool it in one place. For a growing number of important settings, that straightforward route is closed. A hospital cannot send patient records to a central server. A bank cannot export transaction histories to a shared repository. A national digital infrastructure system cannot move citizen data across jurisdictional boundaries. Yet each of these participants would benefit from a model trained on the collective experience of all of them, not just their own slice of it. Federated learning exists to resolve exactly this tension. Instead of moving data to a central model, it moves a model to the data, training locally at each participant and combining only the resulting model updates. This lets many participants benefit from each other's data without any of them ever seeing it.

The combining step is where the central design assumption of federated learning lives. FedAvg~\cite{fedavg}, the foundational algorithm in this space, simply averages the locally trained parameters from each participant, weighted by how much local data each one used. For this to produce a good joint model, the participants' local objectives need to be similar enough that their individually optimal parameters land close together in parameter space. They must be close enough that averaging them lands somewhere reasonable, rather than somewhere that fits none of them well. Real participants are rarely perfectly similar, so a substantial body of work has studied what happens when they are not. This is the well known non-IID problem in federated learning, in which participants' data differ in which labels are common, or in the statistical properties of their features. FedAvg was explicitly designed with this kind of heterogeneity in mind, and its principal descendants, FedProx~\cite{fedprox} and SCAFFOLD~\cite{scaffold}, extend its tolerance further under exactly this kind of label and feature skew. These methods work well within the scope they target, which is clients whose local data differ in distribution while still being samples of a common underlying kind of data, such as images of digits or tabular records sharing one schema. The federated learning literature therefore has good reason to treat non-IID heterogeneity as a largely understood problem.

Graph neural networks introduce a kind of heterogeneity this literature was not built to handle, and it is worth being precise about why. The weights of a graph neural network do not simply parameterize a function over independent feature vectors the way a standard classifier's weights do. Combined with a client's adjacency structure, they define an operator that propagates information along a specific topology, determined by who is connected to whom and how densely. Two clients running the identical shared weight matrix through two structurally different graphs are therefore not in the position of two clients with different label distributions over one shared feature space, evaluating one function on two different inputs. They are closer to asking one set of shared parameters to do two different jobs at once, because the operator those parameters define behaves differently depending on the topology it is applied to. Consider federated clients holding graphs from genuinely different structural domains, such as a citation network, a co-purchase network, a hospital referral network, and a co-authorship network. Averaging their locally trained models is not averaging different samples of one distribution in the way FedAvg was designed to handle. It is averaging the parameters of operators that were fit to do structurally different jobs. This paper investigates empirically whether, and how severely, this actually hurts specific participants in practice, rather than being a concern that sounds plausible but washes out in aggregate.

This question matters beyond its technical interest, because the participants most likely to be harmed by this failure mode are not a random cross-section of a federation. They are disproportionately the participants whose graph structure is atypical relative to everyone else, and structural atypicality correlates in practice with being smaller, newer, or less established. A large hospital network with a mature, well populated referral graph, or a long running citation database, is more likely to resemble a federation's structural majority than a newly onboarded participant still building out a sparser or differently organized graph. If federated training systematically penalizes exactly the participants it was meant to help onboard, that is not a minor optimization inefficiency to be tuned away later. It is a fairness problem embedded in the most widely deployed family of federated learning algorithms. Standard reporting practice makes this easy to miss rather than easy to catch. A federation can show excellent mean accuracy across all participants while one or two structurally atypical participants are being severely and consistently disadvantaged underneath that average, invisible in the headline number a practitioner would actually look at. Section~\ref{sec:exp1} shows this is not a hypothetical worry we are raising for effect. It is exactly what we observe once we look client by client instead of at the mean.

It would be too simple, though, to conclude from this that FedAvg is broken for graphs because it assumes IID data. It does not assume that. Tolerating non-IID data is part of its design brief, and our own results below show FedAvg producing the same directional association we report for FedProx, though not a statistically significant one at the sample size we started with. The more precise and more defensible claim is narrower. Parameter averaging can impose real, client level costs specifically when independently defined graph structures are incompatible with a single shared graph neural network body. None of the correction mechanisms in FedAvg, FedProx, or SCAFFOLD represent or reason about that particular kind of incompatibility, because none of them were built with a notion of client side input structure in mind in the first place. This is worth spelling out mechanism by mechanism, since it is the gap the rest of this paper is built around. The proximal term in FedProx anchors a client's local update to the global model by penalizing parameter space distance. It is a single scalar correction, blind to why a given client's optimum differs from the consensus, whether that reason is label skew or something more structural. The control variates in SCAFFOLD correct for a specific, additive form of client drift, under an implicit assumption that the drift is a consistent bias estimable from a running average. This is well suited to statistical skew, where such a bias is a reasonable model of what is happening, but it is not obviously suited to a client whose entire local loss landscape is shaped by a different graph topology altogether, which is a qualitatively different kind of divergence. Section~\ref{sec:relwork} surveys the surrounding literature in enough detail to make precise what none of it yet provides, covering personalization, clustering, and gradient conflict methods in addition to the three optimizers just discussed.

Having laid out the problem, we should also be explicit about how we approached investigating it, because it shapes everything that follows. We organized the study around a discipline we tried to hold to throughout. Every claim we advance here is one we attempted, and in several cases failed, to disprove. This is a deliberate departure from how results in this area are usually written up, and we adopted it for a specific reason. A correlation or a proposed method that survives a serious, escalating attempt at falsification is more trustworthy than one supported only by the experiments that were originally run to demonstrate it, because the latter can look convincing simply for having stopped looking at the right moment. Under this view, a negative result is informative in its own right rather than something to quietly leave out of the write-up. Such a result might be a plausible mechanism that turns out not to explain the effect it was proposed for, or a mitigation that turns out to be an artifact of an under-tuned baseline rather than a real improvement. This discipline matters especially in a space like federated heterogeneity research, where the space of plausible sounding mechanisms and fixes is large and the temptation to stop at the first one that works is correspondingly strong. Section~\ref{sec:exp4} shows directly why the discipline is not just a rhetorical stance. A comparison against an under-tuned baseline can manufacture an apparently significant, seemingly well validated result out of nothing more than that under-tuning, and only a further, harder test exposed it.

With that groundwork laid, we can state the paper's contributions. It is worth stating up front not just what they are but how they relate to one another, since the structure of the rest of the paper follows directly from this hierarchy rather than from the order the experiments happened to be run in.

\textbf{Primary contribution.} A label free, pre-training degree divergence statistic is associated with which clients are disproportionately harmed by federated graph neural network training. This association is strongest in a small, initial federation, and it remains weaker but still significant when the federation is expanded threefold, including within a single structural domain (Section~\ref{sec:exp1}). A companion spectral divergence statistic showed an even stronger association at small scale but did not replicate at the larger scale, which we trace to a specific, checkable confound rather than leaving unexplained. This is the paper's central, load bearing claim, and everything else in the paper supports, tests, or qualifies it.

\textbf{Secondary contribution.} A strong observational association, however striking, does not by itself mean the measured statistic is causal, and we treat that as a question to be tested rather than assumed. We test it directly with a graph surgery intervention that isolates topology from every other variable, and it does not confirm a simple topology only mechanism (Sections~\ref{sec:exp2} and \ref{sec:exp3}). Motivated by what that null result leaves open, we separately identify and confirm a specific, related mechanistic pattern in how graph convolutional networks aggregate divergent clients' updates, and we show that it too does not fully explain the harm (Section~\ref{sec:exp3b}).

\textbf{Methodological contribution.} Once we had a validated association, the natural next question was whether it could be turned into a working fix. Apparent gains from divergence aware federated learning mechanisms can disappear entirely once compared against a matched strength, structurally blind baseline (Section~\ref{sec:exp4}). This includes a mechanism we designed ourselves, built specifically around the statistics validated in the primary contribution. We think this is a useful caution for how the field evaluates new heterogeneity aware methods generally, independent of the rest of this paper's specific findings.

The remainder of the paper follows this logic in order: Section~\ref{sec:relwork} situates these three contributions against existing federated learning and federated graph learning work; Section~\ref{sec:problem} formalizes the setting and the statistics we use; Section~\ref{sec:setup} describes the federation and training protocol common to every experiment; Sections~\ref{sec:exp1}--\ref{sec:exp4} report the five experiments in the order the contributions above require; and Section~\ref{sec:discussion} onward draws out what the whole picture implies, what it does not yet establish, and what would be needed to establish it.

\section{Related Work}
\label{sec:relwork}

\subsection{Federated Optimization Under Statistical Heterogeneity}

We start where the introduction left off, with the three optimizers whose scope this paper argues is too narrow for structural heterogeneity, examined closely enough to say precisely where that narrowness comes from. FedAvg~\cite{fedavg} averages locally-trained model parameters weighted by local dataset size, after each client performs several local SGD steps. Its implicit assumption is that local objectives, though not identical, are similar enough that their optima lie close together in parameter space, so that averaging moves toward a reasonable joint solution. Under substantial label or feature skew this assumption degrades: local optima drift apart, and naive averaging can converge slowly or to a poor solution, a phenomenon usually called \emph{client drift}. This is the specific failure FedAvg's descendants were built to correct, and understanding how they correct it is what lets us see why the correction does not extend to structural heterogeneity.

FedProx~\cite{fedprox} addresses client drift by adding a proximal term, $\frac{\mu}{2}\lVert w-w^{(t)}\rVert^2$, to each client's local objective, penalizing deviation from the current global model. This bounds how far a client's local update can wander in a single round, trading off local progress for global stability, and it is effective precisely when a single global scalar $\mu$ reasonably captures how much every client's local optimum should be expected to deviate. That last clause turns out to matter a great deal for this paper: our own results in Section~\ref{sec:exp4} show this scalar can be badly mis-set by default choices common in the literature, to the point that an entire mitigation effect we initially attributed to a sophisticated divergence-aware mechanism turned out to be fully recoverable from simply choosing a better $\mu$, uniformly, with no structural awareness at all.

SCAFFOLD~\cite{scaffold} takes a more targeted approach to the same underlying problem. It maintains a control variate at the server and at each client, and uses the difference between them to correct the local gradient direction, explicitly compensating for the component of client drift attributable to a consistent directional bias. This is a more surgical correction than FedProx's blunt distance penalty, and SCAFFOLD is known to converge faster under label skew as a result. But SCAFFOLD's correction is derived and validated under statistical, label-and-feature heterogeneity specifically. We found no treatment, either in the original work or in the broad literature building on it, of whether its control-variate mechanism composes correctly with a further, independent source of client divergence rooted in input structure rather than label distribution. This is not a hypothetical gap: Section~\ref{sec:exp1} and Table~\ref{tab:realfed} show a case where the composition fails outright, with a structural-divergence-based correction combined with SCAFFOLD producing a significant \emph{degradation} rather than an improvement, consistent with the two mechanisms' corrections working at cross purposes rather than reinforcing each other.

Beyond these three foundational methods, a broader family has refined client-drift correction along several axes, without changing its basic scope; it is worth surveying briefly, because one member of this family turns out to be conceptually close to a mechanism we test independently in Section~\ref{sec:exp3b}. FedNova~\cite{fednova} corrects for a specific source of objective inconsistency. Clients that perform different numbers of local steps contribute updates of different effective scale, and naive averaging implicitly over-weights clients that took more steps, so FedNova normalizes each client's contribution by its own local step count before aggregating. FedDyn~\cite{feddyn} adds a dynamic regularizer to each client's local objective, updated every round, so that the fixed points of local and global optimization coincide exactly rather than only approximately, as under FedProx's static penalty. FedOpt~\cite{fedopt} generalizes FedAvg's final aggregation step itself, replacing plain averaging with server-side adaptive optimizers such as FedAdam and FedYogi, treating the sequence of aggregated updates across rounds as a first-class optimization trajectory to be accelerated rather than a single-shot average. All three, like FedAvg, FedProx, and SCAFFOLD before them, are evaluated and motivated on vector-valued data where heterogeneity means differing distributions of the same kind of sample; none represents a client's input as anything other than a distribution to be corrected for, and none reasons about topology as a distinct source of divergence. FedNova's step-normalization principle is where the conceptual overlap with our own work lies, because it corrects for a magnitude imbalance in client updates, which is structurally similar to a mechanism we test directly in Section~\ref{sec:exp3b}. We find, however, that in the specific case of topology-induced magnitude asymmetry under GCN degree-normalization, a magnitude correction of this kind does not reliably explain or resolve the harm. This is a result that the evaluation in FedNova, being scoped to statistical heterogeneity, would not have surfaced.

\subsection{Personalization and Clustering}

A different family of responses to FL heterogeneity abandons the goal of one shared model entirely, on the reasoning that if a single consensus model is the wrong unit of aggregation for a heterogeneous population, the fix should target that unit rather than the aggregation rule applied to it. Personalization methods such as FedPer~\cite{fedper} keep some network layers local to each client and share only a subset, typically the earlier, more generic layers, on the premise that not every part of the model should be forced into agreement across clients. FedRep~\cite{fedrep} formalizes a related split explicitly as a shared, globally-trained representation plus a personalized, locally-trained head, alternating between local head updates and global representation updates. Ditto~\cite{ditto} instead trains a fully personalized model per client, regularized toward a jointly-trained global model rather than replaced by it, explicitly trading off global consistency against local fit as a tunable objective; pFedMe~\cite{pfedme} achieves a similar personalization/globalization trade-off through a Moreau-envelope regularization of the local objective, which admits a cleaner convergence analysis than Ditto's more direct regularization. Clustered federated learning approaches such as IFCA~\cite{ifca} go further still, inferring which clients are similar enough to share a model and maintaining several cluster-specific models rather than one global model, which is the most direct acknowledgment in this literature that a single consensus can be the wrong unit of aggregation when a client population is not homogeneous.

Our own architecture, formalized in Section~\ref{sec:problem}, already adopts a limited form of personalization in this same spirit. Each client keeps a local input projection and output head, sharing only the GNN body between clients. This is a natural fit for graph FL specifically, since it accommodates clients with different feature dimensionalities and label spaces without requiring clustering or a fully personalized model per client. But it is worth being precise about what this family of methods still does not give us, since that gap is what motivates this paper's central question. What none of FedPer, FedRep, Ditto, pFedMe, or IFCA provide is a predictive account of \emph{which} clients need protection and how much of it. Their personalization or clustering decisions are applied uniformly across all clients, or inferred from training dynamics after the fact, rather than diagnosed in advance from a client's own, independently measurable structural properties. Section~\ref{sec:exp1} shows this diagnosis is possible, at least correlationally, for graph-structured clients specifically, which is the piece this literature is missing.

\subsection{Federated Learning on Graphs}

Having surveyed general-purpose FL heterogeneity methods, we now turn to work that addresses graphs specifically, since this is the literature closest to our own setting. Federated graph learning is a fast-moving area, and a recent survey by Fu et al.~\cite{fgmlsurvey} is a useful place to orient within it, because it explicitly separates structural data heterogeneity in federated graph learning from the \emph{statistical} non-IID setting that dominates the classical FL literature just surveyed, which is the same distinction this paper is built around, and it suggests the field is increasingly aware of the gap this paper investigates empirically rather than only taxonomically.

Within that broader space, graph-structured FL splits into two settings with genuinely different core challenges, and it is worth distinguishing them before positioning our own work. In the \emph{subgraph-federated} setting, one large graph is partitioned across clients, and the central difficulty is that edges crossing client boundaries are invisible to any single participant; representative approaches such as FedSage+~\cite{fedsage} address this by training a local model to generate plausible missing neighbor information before message passing, and CNFGNN~\cite{cnfgnn} addresses a related cross-node structural dependency problem for spatio-temporal graph data specifically. This is a different problem from ours. The graph is conceptually one object split by partition boundaries, and heterogeneity arises from those boundaries, not from clients belonging to genuinely different structural domains in the first place.

The \emph{cross-domain} setting, where each client holds an independently-defined graph rather than a fragment of one shared graph, is the setting closer to the federation we study, so it is where we position our contribution most directly. Federated graph classification over non-IID graphs, as studied by approaches such as GCFL~\cite{gcfl}, explicitly clusters clients by gradient or structural similarity before aggregating within clusters, again acknowledging, as the personalization literature does, that heterogeneous graph populations may not tolerate a single shared model well; personalized subgraph federated learning, as studied by Baek et al.~\cite{baek2023}, pursues a related idea, training client-specific graph representations rather than forcing full parameter sharing. Closer still, Tan et al.~\cite{fedstruct} address structural non-IIDness directly, sharing structural knowledge, rather than only parameters, across clients to mitigate the effect of differing graph topology; this is the closest prior work we are aware of to the specific problem this paper studies. It differs from our approach in two respects worth naming precisely, since both are exactly what this paper adds. Their evaluation reports aggregate accuracy improvements without an accompanying causal analysis of why, or for which clients, structural sharing helps. That is the question Section~\ref{sec:exp3} takes on directly. And their comparison does not include the baseline-tuning control we show, in Section~\ref{sec:exp4}, to be necessary before attributing any gain to structural awareness specifically, rather than to a baseline that simply was not tuned as carefully as the proposed method. Recent benchmarking efforts, including OpenFGL~\cite{openfgl} and FedGraph~\cite{fedgraph}, standardize evaluation protocols and datasets across this growing literature, which we read as further evidence that the field is actively moving toward the kind of evaluation rigor Section~\ref{sec:exp4} argues for, even where it is not yet standard practice. Taken together, as far as we can tell, this literature evaluates methods primarily on aggregate or mean accuracy across clients; we did not find a validated, label-free, \emph{a priori} predictor of which specific clients will be harmed by federation reported elsewhere, nor negative or inconclusive results reported with the granularity we report in Sections~\ref{sec:exp2}--\ref{sec:exp4}. Our contribution is complementary to this line of work rather than competing with it directly. Rather than proposing another clustering or aggregation rule, we ask what specific, measurable property of a client's graph is associated with harm in the first place, independent of which correction mechanism is eventually applied on top of that diagnosis.

\subsection{Gradient Conflict and Robust Aggregation}

One more family of methods is relevant here, and it is the family our own earlier baseline mechanism belongs to, which is why we cover it last before turning to our own design. Outside the FL literature specifically, gradient-surgery methods from multi-task learning, such as PCGrad~\cite{pcgrad}, detect pairwise conflicts between task gradients and project out the conflicting component before combining them, on the premise that na\"ively summing conflicting gradients can actively harm both tasks involved. GradNorm~\cite{gradnorm} takes a related but distinct approach, equalizing the \emph{magnitude} of gradients across tasks by dynamically re-weighting each task's loss so that no single task's gradient dominates the shared update purely by having a larger scale. This is a magnitude-based correction conceptually related to the mechanistic hypothesis we test in Section~\ref{sec:exp3b}. CAGrad~\cite{cagrad} generalizes gradient surgery further still, seeking an update direction that maximizes worst-case improvement across all tasks simultaneously subject to a bound on deviation from the average gradient, rather than resolving conflicts pairwise as PCGrad does. Model-contrastive approaches such as MOON~\cite{moon} take a different tack entirely, regularizing each client's local representation to stay close to the previous global representation and away from its own previous local representation, using a contrastive loss computed entirely in representation space rather than parameter space. Robust and Byzantine-aware aggregation rules detect and down-weight or discard client updates that are statistical outliers relative to the rest. This family includes coordinate-wise trimming or median-based estimators~\cite{byzrobust}, geometric-median-based robust aggregation~\cite{rfa}, and similarity-based schemes such as Krum~\cite{krum}. These methods were originally motivated by adversarial or corrupted clients, not by organically differing but entirely legitimate structural regimes of the kind we study.

The mechanism we use as a baseline in Section~\ref{sec:exp4}, which we refer to throughout this paper simply as the \emph{carrier mechanism} to avoid implying it is published prior work rather than our own earlier design, belongs to this same family. Concretely, it maintains a single evolving reference direction in parameter space, which we call a carrier by analogy with a reference signal that other signals are compared and aligned against. This direction is computed as a running average of the update directions of clients that agree with it above a threshold in a given round. Each client's raw update is then processed in three steps. It is soft-weighted by its cosine alignment with the current carrier direction; projected onto a low-dimensional subspace spanned by the recent history of well-aligned updates, refreshed periodically; and clipped to a maximum norm. The resulting post-processed vector's norm, relative to the other clients' post-processed norms, becomes a per-client scale factor that reweights the ordinary federated average, leaving each client's original update direction untouched and only rescaling its contribution. This composition of three regulation steps is, to our knowledge, novel to our own earlier work and not otherwise published; we include it here purely as an internal baseline for comparison against the newer mechanisms we design, not as prior literature we are building on.

What unites every method surveyed in this subsection, the carrier mechanism included, is a shared structural weakness worth naming explicitly, because it is the direct motivation for what we try instead in Section~\ref{sec:exp4}. All of them regulate \emph{after the fact}, at the level of the aggregated update or representation, using a signal computed from the very training run they are trying to correct. None of them uses an independently-computed, pre-training, label-free structural statistic of the kind we validate in Section~\ref{sec:exp1}. This is the direct motivation for the divergence-calibrated mechanisms we design and test in Section~\ref{sec:exp4}: rather than inferring conflict from parameter-space geometry during training, as every method above does, we ask whether a statistic computable \emph{before} training even starts can calibrate the correction directly, in advance. Our results turn out to be a mixed answer to that question. The diagnostic statistic itself is validated, but the calibrated mechanisms we built from it did not, in the end, outperform a properly-tuned, structurally-blind baseline (Section~\ref{sec:exp4}). Once we reach that section we discuss why this particular family of after-the-fact regulation methods, our own carrier mechanism included, may be harder to validate rigorously than it first appears.

\subsection{Positioning and Gap}

Synthesizing the four strands surveyed above, three questions motivate this paper's experiments directly, and we deliberately phrase them as open questions rather than novelty claims, since federated graph learning is a fast-moving literature and we cannot rule out that some corner of it already addresses one of these in a way we have not seen. First: can the client-level harm from federation be identified \emph{before} training starts, from a measurable property of a client's own graph, with no labels involved at all? Personalization and clustering methods, as we have seen, respond to heterogeneity once it manifests; they do not, as far as we can tell, diagnose it in advance. Second: when a heterogeneity-aware or gradient-conflict mechanism is evaluated against a baseline, has that baseline's own hyperparameters been searched with comparable care to the proposed method? We did not find this control applied in the comparable work we reviewed, and Section~\ref{sec:exp4} shows it is decisive rather than a formality, and this applies to our own earlier carrier mechanism as well. Third: is the structural-heterogeneity-harm relationship, wherever it is reported in this literature, subjected to an explicit causal test, rather than left as a correlational or purely algorithmic, does-it-improve-accuracy finding? We report such a test honestly as inconclusive in Section~\ref{sec:exp3}, and we are not aware of a comparable test elsewhere in this specific literature. These three questions, taken together, are the direct motivation for the five experiments that follow.

\section{Problem Formulation}
\label{sec:problem}

Before we can test any of the three questions just raised, we need a precise, shared vocabulary for what a client's structure is, what harm means for that client, and how we will measure divergence between clients without reference to labels. This section establishes that vocabulary; the experiments beginning in Section~\ref{sec:exp1} apply it directly.

\subsection{Federation and Architecture}
We consider $K$ clients, each holding a graph $G_k = (V_k, E_k)$ with node features $X_k \in \mathbb{R}^{n_k \times d_k}$ and labels $y_k$. Critically, and unlike the standard FL setting surveyed in Section~\ref{sec:relwork}, $d_k$ (feature dimensionality) and the structural properties of $G_k$ may differ arbitrarily across clients. Clients need not be samples from a common domain at all, which is precisely the departure from ordinary non-IID FL this paper studies. To accommodate that departure architecturally, each client's model factors as
\[
f_k(X_k, A_k) = O_k\big(\,\Phi_\theta(P_k(X_k), A_k)\,\big),
\]
where $P_k: \mathbb{R}^{d_k}\to\mathbb{R}^h$ is a client-local input projection, $O_k: \mathbb{R}^h \to \mathbb{R}^{C_k}$ is a client-local output head accommodating different label spaces $C_k$, and $\Phi_\theta$ is an $L$-layer GNN body of fixed hidden width $h$, \emph{shared and federated} across all clients. This split is what lets clients with different $d_k$ and $C_k$ participate in the same federation at all. Only $\theta$, the body's parameters, is communicated and aggregated, while $P_k$ and $O_k$ remain local to each client throughout training, in the same spirit as the personalization methods discussed in Section~\ref{sec:relwork}. We test $\Phi_\theta$ instantiated as both GCN~\cite{gcn} (symmetric-normalized adjacency, $H^{(l+1)} = \mathrm{ReLU}(\tilde A H^{(l)} W^{(l)})$) and GraphSAGE~\cite{graphsage} (mean-neighbor aggregation with separate self/neighbor transforms), so that our findings are not an artifact of one specific aggregation rule.

\subsection{Negative Transfer}
With the architecture fixed, we can now define what it means for federation to harm a client, borrowing terminology from the transfer-learning literature, where \emph{negative transfer} denotes a case in which combining information from an auxiliary source actively degrades performance relative to not using that source at all~\cite{negtransfer}. For client $k$, we define
\[
\Delta_k \;=\; \mathrm{acc}_k^{\text{local-only}} - \mathrm{acc}_k^{\text{federated}},
\]
the accuracy $k$ would achieve training $\Phi_\theta, P_k, O_k$ alone on its own graph for a matched compute budget, minus the accuracy it actually achieves as a federation participant under a given aggregation rule. $\Delta_k > 0$ indicates federation actively hurt client $k$ relative to simply going it alone. This is the quantity every subsequent experiment in this paper is ultimately trying to explain, predict, or reduce.

\subsection{Structural Divergence Metrics}
Explaining $\Delta_k$ is only useful if we can measure a candidate cause of it without already knowing $\Delta_k$ itself; otherwise we would have no way to use the measure predictively, only descriptively after the fact. We therefore characterize client $k$'s structural divergence from the rest of the federation with two statistics, deliberately constructed to require no label information and to be computable purely from a client's graph structure, before any training occurs:
\begin{itemize}
\item \textbf{Degree divergence}. This is the Wasserstein-1 distance~\cite{emd,villani} between the size-normalized degree sequence of client $k$, written $\{d_i/n_k\}$ for $i \in V_k$, and the pooled, size-normalized degree sequence of all other clients.
\item \textbf{Spectral divergence}: the Wasserstein-1 distance~\cite{emd,villani} between the top-$m$ eigenvalues of $k$'s symmetrically-normalized adjacency operator $\tilde A_k = D_k^{-1/2}(A_k+I)D_k^{-1/2}$ and the pooled top-$m$ eigenvalues of all other clients' operators.
\end{itemize}
We additionally define, and later test as candidate explanations in their own right, a \textbf{homophily gap} (absolute difference in edge-label-homophily from the federation mean) and a \textbf{feature-dimensionality divergence} $|\log_2(d_k/\bar d_{-k})|$, signed to allow for asymmetric effects in either direction. With these definitions fixed, we turn next to the concrete federation we built to test them.

\section{Experimental Setup}
\label{sec:setup}

\subsection{Real Federation}
To test whether the statistics just defined actually track real-world harm, we needed a federation with genuine, uncontrived structural heterogeneity rather than a purely synthetic construction designed to make our own hypothesis look good. We therefore constructed a six-client federation deliberately spanning real and synthetic structural domains (Table~\ref{tab:clients}): three real citation graphs (Cora, CiteSeer, PubMed, originally introduced by Sen et al.~\cite{sen2008} and loaded via the standard Planetoid split~\cite{yang2016}) and three synthetic stochastic block model (SBM)~\cite{sbm} graphs standing in for denser, differently-organized domains such as co-purchase and co-authorship networks, with node features generated as class-centroid-plus-Gaussian-noise so the classification task is genuinely learnable from the synthetic topology and features jointly, rather than trivially easy or impossible. The result is substantial structural heterogeneity by construction. Mean degree spans a nearly 9$\times$ range (2.77--24.75) and feature dimensionality spans a 37$\times$ range (100--3703) across the six clients.

\begin{table*}[t]
\caption{The six-client real federation. Structural heterogeneity spans both real and synthetic domains.}
\label{tab:clients}
\begin{tabular}{lllrrrrr}
\toprule
Client & Role & $n$ & $d$ & $C$ & Mean deg. & Density \\
\midrule
cora & citation & 2708 & 1433 & 7 & 3.90 & 0.0014 \\
citeseer & citation & 3327 & 3703 & 6 & 2.77 & 0.0008 \\
pubmed & citation & 19717 & 500 & 3 & 4.50 & 0.0002 \\
sbm-dense-a & co-purchase proxy & 1800 & 100 & 5 & 24.75 & 0.0137 \\
sbm-dense-b & co-purchase proxy & 2400 & 100 & 8 & 17.42 & 0.0073 \\
sbm-coauthor & co-authorship proxy & 1500 & 100 & 4 & 9.18 & 0.0061 \\
\bottomrule
\end{tabular}
\end{table*}

\subsection{Training Protocol}
Every experiment in this paper, unless a specific section notes a deliberate deviation, shares one training protocol, so that differences we report across experiments reflect the variable under study rather than incidental differences in setup. Unless otherwise noted: hidden width $h=32$, $L=2$ GNN layers, Adam~\cite{adam} local optimizer (learning rate 0.01), 5 local steps per round, 50 federated rounds, 3 seeds per configuration, extended to 10--24 seeds where a specific section notes that statistical power required it. Client aggregation weights are proportional to local training-set size unless a mitigation explicitly overrides this, which we always flag when it happens. All reported accuracies are test-set accuracies on the client's own held-out split. With the setup fixed, we turn to the first and central experiment, which asks whether the structural divergence statistics just defined actually track real harm in this federation.

\section{Experiment 1: Real-Federation Validation}
\label{sec:exp1}

\subsection{The Original Six-Client Result}

\begin{figure*}[t]
\centering
\includegraphics[width=0.85\linewidth]{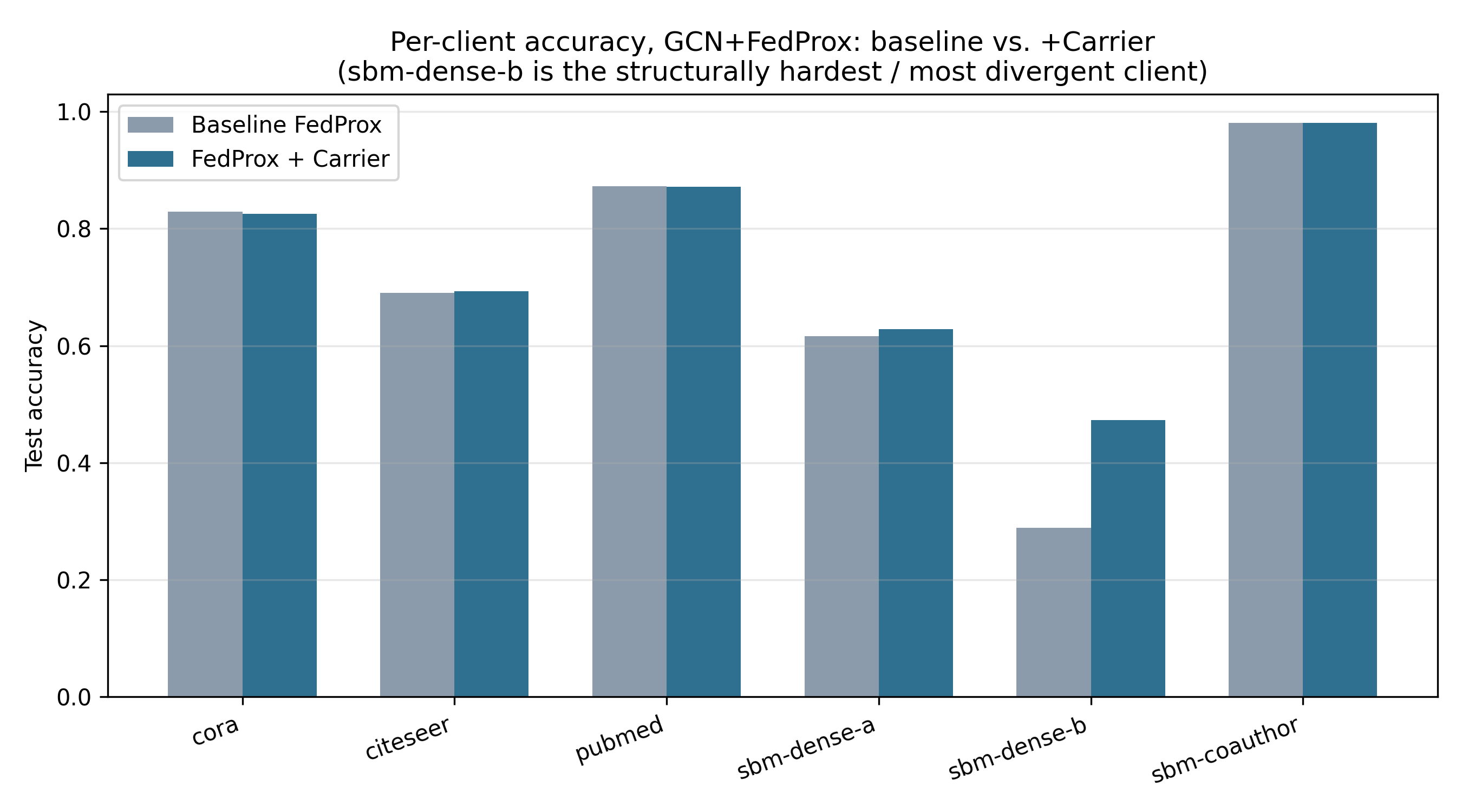}
\caption{Per-client accuracy under GCN+FedProx, baseline vs.\ the gradient-conflict regulation mechanism. The mechanism's benefit is concentrated almost entirely on sbm-dense-b, the structurally most divergent client, and leaves the well-aligned majority essentially untouched. This is a fairness and equity effect that is invisible in the mean-accuracy figure alone.}
\label{fig:perclient}
\end{figure*}

The most direct test of our central hypothesis is simply to train the federation described above and see whether harm is uneven across clients, and if so, whether our structural statistics predict which clients bear it. Table~\ref{tab:realfed} summarizes per-optimizer aggregate results across the six-client federation for both architectures. The first thing to notice is that negative transfer was substantial and highly uneven across clients, not a small, evenly-distributed tax on everyone. The accuracy of sbm-dense-b under plain FedAvg fell to 22--30\% depending on optimizer, versus 50\% achievable training alone. This is a loss of over half its achievable accuracy purely from participating in federation (Figure~\ref{fig:perclient}).

\begin{table*}[t]
\caption{Mean accuracy across the six-client federation, by architecture and optimizer, with (+carrier) and without the carrier mechanism, an earlier gradient-conflict regulation mechanism of our own design (Section~\ref{sec:exp4}). Values are means over 3 seeds.}
\label{tab:realfed}
\begin{tabular}{llrr}
\toprule
Architecture & Optimizer & Baseline & +Carrier \\
\midrule
GCN & FedAvg & 0.745 & 0.747 \\
GCN & FedSGD & 0.551 & 0.547 \\
GCN & FedProx & 0.713 & \textbf{0.745}$^*$ \\
GCN & SCAFFOLD & 0.719 & 0.724 \\
GraphSAGE & FedAvg & 0.803 & 0.794 \\
GraphSAGE & FedSGD & 0.600 & 0.590 \\
GraphSAGE & FedProx & 0.813 & 0.810 \\
GraphSAGE & SCAFFOLD & 0.861 & 0.819$^\dagger$ \\
\bottomrule
\end{tabular}
\\[2pt]
\footnotesize $^*$Significant improvement, GCN+FedProx: $p=0.024$. $^\dagger$Significant degradation, SAGE+SCAFFOLD: $p=0.013$. All other differences not significant at $n=3$.
\end{table*}

\begin{figure}[t]
\centering
\includegraphics[width=\linewidth]{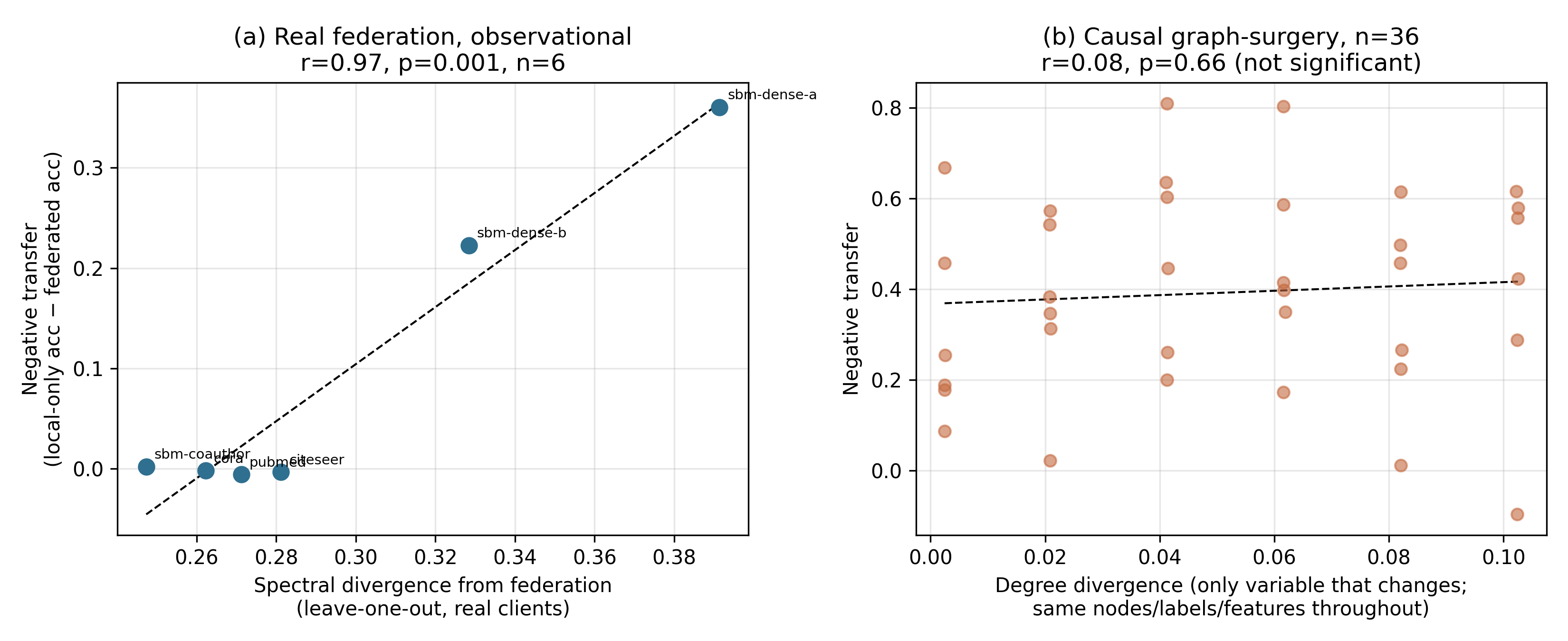}
\caption{(a) Spectral divergence from the rest of the real federation (leave-one-out, label-free) correlates very strongly with negative transfer under FedProx, in the original six-client federation. (b) A true causal intervention, holding nodes, labels, and features identical throughout and interpolating only the edge-generating probabilities between a high-divergence and a majority-matched setting, shows no significant relationship, replicated at $n{=}800$ and $n{=}3000$ node scales.}
\label{fig:corr1}
\end{figure}

Having established that harm is real and uneven, the next question is whether our two structural statistics actually track it, since an uneven pattern of harm that our statistics cannot predict would be a curiosity, not a usable diagnostic. Degree and spectral divergence, computed leave-one-out with no reference to any accuracy number, were strongly associated with this pattern of harm, at a strength uncommon in this literature (Figure~\ref{fig:corr1}a): under FedProx, degree divergence gave $r=0.90$ ($p=0.015$, $n=6$) and spectral divergence gave $r=0.97$ ($p=0.001$). Under plain FedAvg the same two statistics gave $r=0.63$ and $r=0.77$, showing the same direction and a comparable magnitude but not reaching statistical significance at $n=6$. We read this as FedAvg showing the same underlying association at lower statistical confidence rather than a genuinely different relationship, since six clients is not enough to resolve a moderate correlation from noise even when the point estimate looks similar to FedProx's. To check that this association was specific to the statistics we had chosen, rather than any vaguely ``structural'' quantity we might have tested, we also tried a homophily-gap statistic in a separate controlled sweep (Section~\ref{sec:exp2}); it showed no association once density was controlled, which is itself informative. It tells us the effect is specific to density and spectral structure, not a generic notion that more heterogeneous is worse, which any structural statistic would pick up.

It is worth pausing here to be precise about what kind of claim six data points can and cannot support, because the distinction matters for how the rest of this paper should be read. The statistics are genuinely computable before training. That part is not in question, and it is what makes them potentially useful in practice regardless of what follows. Whether the association would generalize beyond these six specific clients was, at this point in the study, a genuinely open question that $n=6$ could not settle on its own; an association this strong at small $n$ is a striking finding worth reporting, but it demonstrates predictive availability, meaning the statistic exists and can be computed in advance, rather than validated predictive generalization at any larger scale. We took this distinction seriously enough to act on it, and Section~\ref{sec:exp1b} reports what happened when we tried to settle the open question directly, by testing whether the association survives a substantially larger and more diverse federation.

\begin{figure}[t]
\centering
\includegraphics[width=0.85\linewidth]{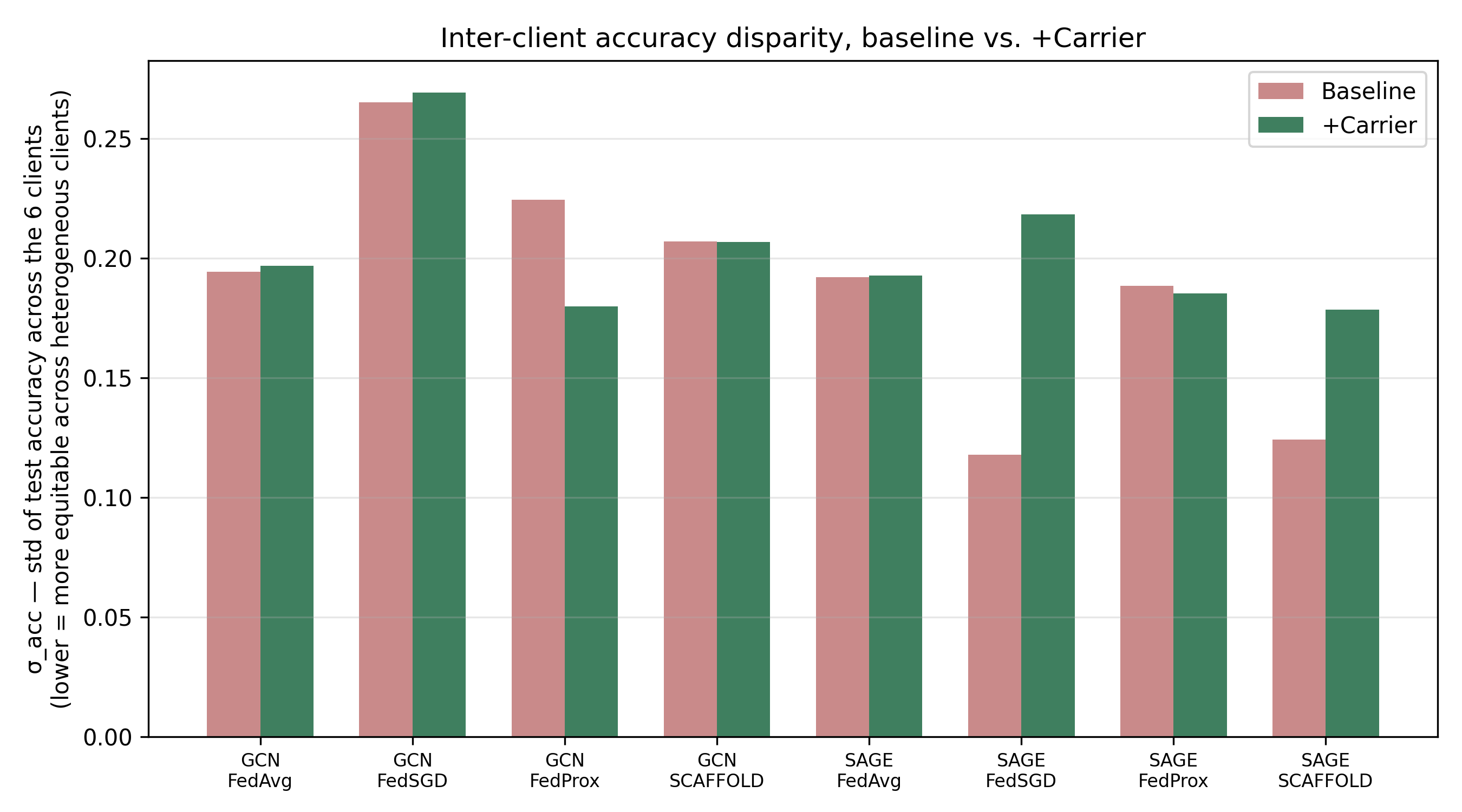}
\caption{Inter-client accuracy disparity ($\sigma_{\text{acc}}$) across all tested optimizer/architecture/mitigation combinations on the six-client real federation. Lower is more equitable across the six structurally heterogeneous clients.}
\label{fig:disparity}
\end{figure}

Before turning to that expansion, one further lens on the same six-client result is worth reporting, because it reveals something mean accuracy hides. Inter-client disparity ($\sigma_{\text{acc}}$, the standard deviation of accuracy across the six clients) shows which clients bear the cost of federation in a way a single mean-accuracy number cannot. Figure~\ref{fig:disparity} shows $\sigma_{\text{acc}}$ across all eight optimizer/architecture/mitigation combinations; the mechanism that produced the one significant mean-accuracy improvement (GCN+FedProx+carrier mechanism) also reduced disparity (0.224$\to$0.180), consistent with the mechanism concentrating its effect on the worst-off client specifically rather than improving everyone uniformly. This is an equity effect, not just an accuracy effect.

\subsection{Expanding to Twenty Clients: What Changed, and Why}
\label{sec:exp1b}

Six clients, however striking their correlation, cannot on their own distinguish a genuine, generalizable dose-response relationship from an association that happens to hold on one small, particular sample. The most direct way to tell these two possibilities apart was not to argue about it in the abstract but to add more clients and see, concretely, whether the association survived contact with a larger and more varied population. We therefore extended the federation from six clients to twenty, keeping the original six entirely unchanged and adding fourteen new synthetic clients generated from the same stochastic-block-model procedure as the original three synthetic clients, but spanning a deliberately wide and continuous range of density and community structure. Mean degree ran from roughly 3 to 25 across the full set of twenty, rather than showing the six original clients' more clustered spread at the extremes of that range. We recomputed leave-one-out degree and spectral divergence for all twenty clients against this larger pool, retrained the full twenty-client federation under FedAvg and FedProx (3 seeds each, same architecture and training protocol as Section~\ref{sec:setup}), and retrained each client's local-only reference under the same protocol.

The association did not survive at its original strength, and, importantly, it did not survive evenly across the two statistics. That detail turns out to be the key to understanding what happened. Degree divergence remained significant but became substantially weaker: $r=0.53$ ($p=0.016$) under FedAvg and $r=0.49$ ($p=0.027$) under FedProx, down from $r=0.90$ and a comparable point estimate at $n=6$. Spectral divergence, the stronger of the two original predictors, lost its association entirely: $r=-0.05$ ($p=0.85$) under FedAvg and $r=0.15$ ($p=0.53$) under FedProx, which is indistinguishable from no relationship at all. Rank-based (Spearman) correlations for degree divergence were also not significant at this sample size ($r=0.26$--$0.33$, $p=0.15$--$0.27$), which tells us something further. Even the surviving Pearson association is not yet a robust, monotonic relationship across the full range of clients, and may be sensitive to a small number of high-divergence points rather than reflecting a smooth dose-response across the whole distribution.

\begin{figure*}[t]
\centering
\includegraphics[width=\linewidth]{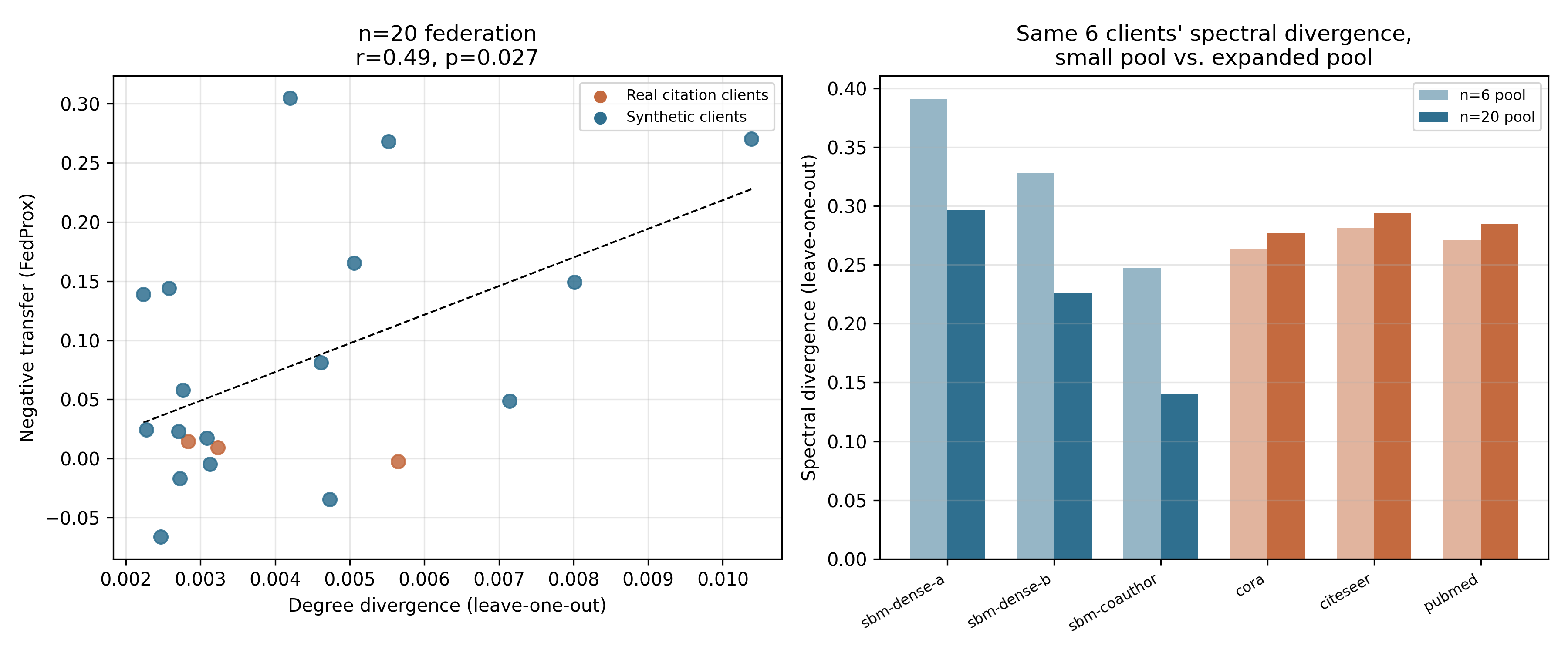}
\caption{(a) The twenty-client federation, degree divergence vs.\ negative transfer under FedProx, colored by domain. The association survives ($r=0.49$) but is far noisier than the original six-client result. (b) The same six original clients' spectral divergence, computed against the original six-client pool vs.\ the expanded twenty-client pool. The three synthetic clients' apparent divergence drops substantially once more same-domain synthetic clients join the pool; the three real citation clients' divergence is essentially unchanged, because they remain a small minority regardless of pool size.}
\label{fig:n20}
\end{figure*}

This is a real reduction in evidential strength, and we do not want to understate it by moving past it quickly. But this is not simply a case of the effect getting weaker for reasons we cannot identify. There is a specific, checkable explanation for the pattern we just described, and it points to something worth knowing about this whole class of diagnostic statistic, not just about our own. Figure~\ref{fig:n20}b shows what happened to the original six clients' spectral divergence values as the pool grew. The three synthetic clients' divergence dropped substantially (sbm-dense-a: $0.391\to0.296$; sbm-dense-b: $0.328\to0.226$; sbm-coauthor: $0.247\to0.140$), while the three real citation clients' divergence barely moved (cora: $0.263\to0.277$; citeseer: $0.281\to0.294$; pubmed: $0.271\to0.285$). The reason becomes clear once we recall exactly what a leave-one-out statistic measures. It measures divergence from whatever pool happens to be present at the time, not from some fixed, external notion of typicality. When we added fourteen more synthetic clients, the original three synthetic clients stopped looking unusual, because they now had plenty of structurally similar neighbors to be compared against; the real citation clients, still a small minority at three out of twenty, looked just as different from the pool as they always had. Put together, this means that in the original six-client federation, spectral divergence was working almost as a detector of whether a client was real or synthetic. Three clients came from one generative process and three from another, with the metric cleanly separating them, and that separation happening to align with which clients suffered the most harm. That is a coherent explanation both for why the original correlation was so strong, and for why it turned out to be fragile once the domain balance changed.

If this diagnosis is correct, it makes a specific, testable prediction. A statistic that only tracks domain identity should show little association \emph{within} a single domain, once domain membership is no longer available as a confound to exploit, whereas a statistic that captures something more structurally real should still show an association even when every client being compared was generated by the same process. We tested this prediction directly by restricting the correlation to the seventeen synthetic clients only, removing the three real clients and, with them, any domain contrast at all. Within this synthetic-only subset, degree divergence remained associated with harm at a similar or even slightly stronger level than in the full twenty-client set ($r=0.56$, $p=0.019$ under FedAvg; $r=0.51$, $p=0.035$ under FedProx), while spectral divergence was not significant under FedAvg ($r=0.26$, $p=0.31$) and only marginally significant under FedProx ($r=0.50$, $p=0.041$). The prediction held. The signal in degree divergence survives with the domain confound removed, and the signal in spectral divergence mostly does not.

We read this as convergent with, rather than contradicting, the rest of this paper's findings, which is reassuring given how much the picture changed between six and twenty clients. Section~\ref{sec:exp3b} independently identifies a mechanism tied specifically to degree and GCN's degree-normalization, not to a more holistic notion of structural dissimilarity; Section~\ref{sec:exp2}'s density sweep, run entirely independently of this expanded validation and before we knew its result, also isolates density as the property that matters and finds homophily-driven divergence does not predict harm. The twenty-client expansion adds a third, independently-motivated line pointing the same way. Degree divergence carries a real signal that survives removing an entire confound, while spectral divergence's apparent power at small $n$ looks, in retrospect, substantially attributable to it doubling as a real-vs-synthetic domain marker in a federation where domain and harm happened to be entangled by construction.

Given all of this, we revise our claim accordingly rather than holding onto the stronger, more striking version. The paper's primary contribution is now more precisely stated. Degree divergence, specifically, is associated with federated GNN negative transfer, at a real but modest strength that survives a roughly threefold increase in sample size and the removal of a specific domain confound; spectral divergence's role is less clear than our initial six-client result suggested, and we no longer present it as an equally strong predictor. We discuss what this implies for the diagnostic's practical use once we reach Section~\ref{sec:discussion}. With the observational case made and qualified, the next question is whether any of this is causal. That is a question the six-client and twenty-client federations, being observational by construction, cannot answer on their own.

\section{Experiment 2: Controlled Synthetic Dose-Response}
\label{sec:exp2}

The real-federation result in Section~\ref{sec:exp1} is observational in a specific sense worth spelling out. The clients differ in structure, domain, and feature dimensionality all at once, so a strong correlation cannot by itself tell us which of these factors, or which combination of them, actually drives the harm. Untangling that requires a design where we control everything except the one variable we want to test. We therefore constructed a controlled synthetic environment: a fixed, structurally-homogeneous four-client ``majority'' federation, using identical SBM generative parameters with only the random seed differing between the four, against which a single held-out ``test'' client's structural properties were swept systematically along independent axes, holding everything else fixed by construction. This design lets us ask, one variable at a time, exactly which structural property is doing the work.

\textbf{Density.} Sweeping the test client's edge density, with both within-community and between-community probabilities scaled together so that their ratio, and hence homophily, stayed fixed, produced a clean, statistically significant dose-response. Degree divergence gave $r=0.50$ ($p=0.034$, $n=18$), spectral divergence gave $r=0.47$--$0.49$ ($p\approx0.05$), consistent in both direction and magnitude with Experiment 1. This is the first independent confirmation that density specifically, not structural heterogeneity in some vaguer sense, is doing the work.

\textbf{Homophily.} Having confirmed density matters, the natural next question is whether homophily, the other obvious candidate structural property, matters too, independent of density. Sweeping the within/between edge-probability ratio while holding overall density fixed produced \emph{no} significant relationship with negative transfer ($r=-0.28$ to $-0.37$, $p>0.12$, $n=18$). This null result sharpens the claim in a useful direction. Density is associated with harm and homophily divergence at fixed density is not, which rules out a generic ``more heterogeneous is worse'' story in favor of something more specific to density and connectivity.

\begin{figure*}[t]
\centering
\includegraphics[width=0.8\linewidth]{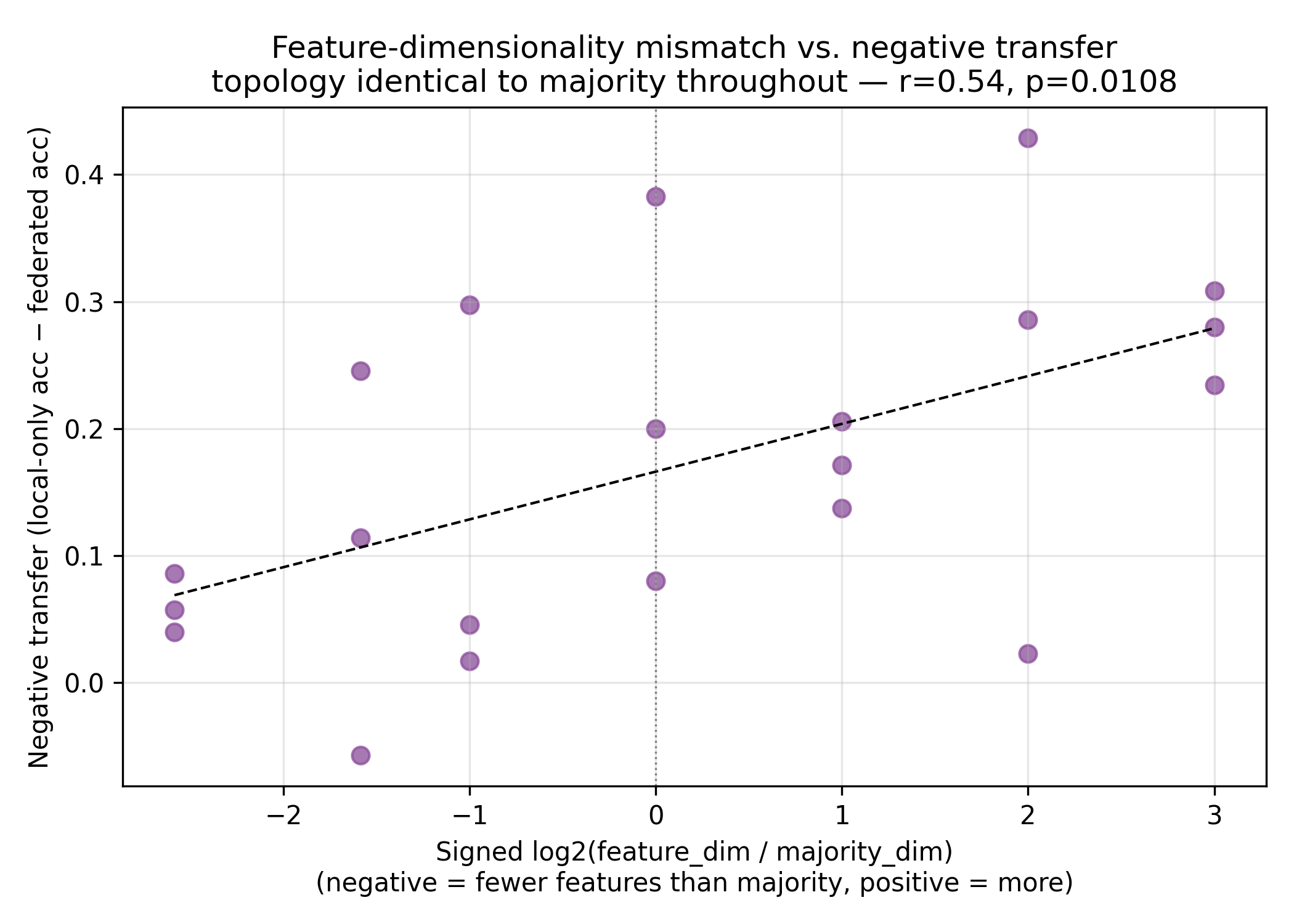}
\caption{Feature-dimensionality mismatch (topology held identical to the majority throughout) predicts negative transfer asymmetrically. Excess dimensionality relative to the federation is costly, while deficit dimensionality is largely benign.}
\label{fig:featdim}
\end{figure*}

\textbf{Feature dimensionality.} A different candidate explanation is that harm comes from feature-space mismatch rather than topology at all, so we tested this too, holding topology identical to the majority and sweeping only the test client's input feature dimensionality. This revealed an asymmetric effect worth reporting on its own terms. A signed log-ratio metric correlated significantly with negative transfer ($r=0.54$, $p=0.011$, $n=21$), driven almost entirely by clients with \emph{more} features than the majority (up to 27\% negative transfer at $4\times$ the majority's dimensionality), while clients with fewer features than the majority showed little effect ($\sim$6\%). Figure~\ref{fig:featdim} shows this relationship. This tells us feature-space mismatch is a real, independent contributor to harm, separate from the degree effect, though as Section~\ref{sec:exp1} already noted, we could not validate this specific statistic independently in the real federation, where it was confounded with client role.

\textbf{Community granularity (confounded, excluded from causal claims).} A fourth sweep, varying the number of communities and hence output classes in the test client's graph, initially appeared to show a strong effect, but following it up revealed it was confounded twice over, in a way worth reporting as a cautionary methodological note rather than quietly dropping. First, a model-capacity artifact was at work, because a small hidden-width GCN cannot cleanly separate 32-way classification regardless of data volume, producing a floor effect at high $k$ that has nothing to do with structural divergence. After correcting model capacity, a deeper and uncorrectable confound remained. Varying community count also varies the test client's output space (2-way vs.\ 32-way classification) relative to the fixed 5-way majority, conflating topological granularity with task-cardinality mismatch, which is a fundamentally different kind of heterogeneity than the one we set out to study. We exclude this axis from our causal claims and report it here as a design lesson worth keeping in mind for future work in this space. Not every plausible-sounding structural sweep is actually isolating structure, and it is worth checking before trusting the result.

Taken together, these four sweeps narrow the candidate explanations for negative transfer considerably. Density and feature-dimensionality mismatch are real, independent contributors, homophily divergence at fixed density is not; and community granularity cannot be cleanly tested with this design. But even the density and feature-dimensionality results, clean as they look, share a limitation the real-federation result also had. Each sweep independently re-sampled node features and labels at every point along the density axis, which leaves open the possibility that density and some incidental, unmeasured feature separability were confounded across the sweep rather than being genuinely decoupled. Correlation, however strong and however carefully controlled, is still not evidence of causation on its own~\cite{holland1986}. Section~\ref{sec:exp3} addresses this limitation directly.

\section{Experiment 3: Causal Graph Surgery}
\label{sec:exp3}

Establishing causation, rather than a merely well-controlled correlation, requires an intervention that manipulates the variable of interest while holding every other factor fixed. The factors must be fixed not just in expectation across a sweep, as in Section~\ref{sec:exp2}, but bit-for-bit in the specific instance being tested. We designed exactly this kind of intervention. One draw of nodes, community assignment, and features was fixed \emph{once}; only the edge-generating probabilities were then interpolated in six steps between a high-divergence (``native'') and majority-matched (``target'') setting, with everything else held bit-for-bit identical across the interpolation, so that any change in outcome could only be attributed to the change in topology. We replicated this at two graph scales, $n=800$ (20 independent edge draws per step, to characterize a first, necessarily noisy result) and $n=3000$ (6 draws per step, specifically to test whether the noise at the smaller scale was a small-graph SBM sampling artifact rather than a real absence of effect).

At neither scale did degree divergence show a significant relationship with negative transfer under this genuine intervention ($r=0.07$, $p=0.47$, $n=120$ pooled at $n{=}800$; $r=0.08$, $p=0.66$, $n=36$ pooled at $n{=}3000$; Figure~\ref{fig:corr1}b). We want to be direct about what this means rather than explain it away after the fact. The correlational relationship documented in Sections~\ref{sec:exp1}--\ref{sec:exp2} is real, large, and reproducible on its own terms, but a directly isolated causal test of degree divergence as the mechanism does not confirm it as the cause. This does not mean the association is spurious, since the twenty-client replication in Section~\ref{sec:exp1b}, run independently of this causal test, still found it. It does mean we cannot yet claim to know why the association holds. That open question motivates the next experiment.

\section{Experiment 4: A Highly Consistent but Inert Mechanism}
\label{sec:exp3b}

If topology alone, tested as directly as Section~\ref{sec:exp3} tested it, does not explain the harm, a natural next step is to ask whether a more specific, architecturally-grounded mechanism might, even if the general causal test came back null. We tested one such hypothesis, motivated by the architecture of GCN itself rather than by structural divergence in the abstract. We hypothesized that degree-normalization ($\tilde A = D^{-1/2}(A+I)D^{-1/2}$) systematically shrinks the magnitude of a structurally divergent, denser client's shared-body parameter update, such that its contribution is simply out-voted in a magnitude-weighted federated average whatever its direction, independent of whether that contribution was actually pointing the right way.

\begin{figure*}[t]
\centering
\includegraphics[width=\linewidth]{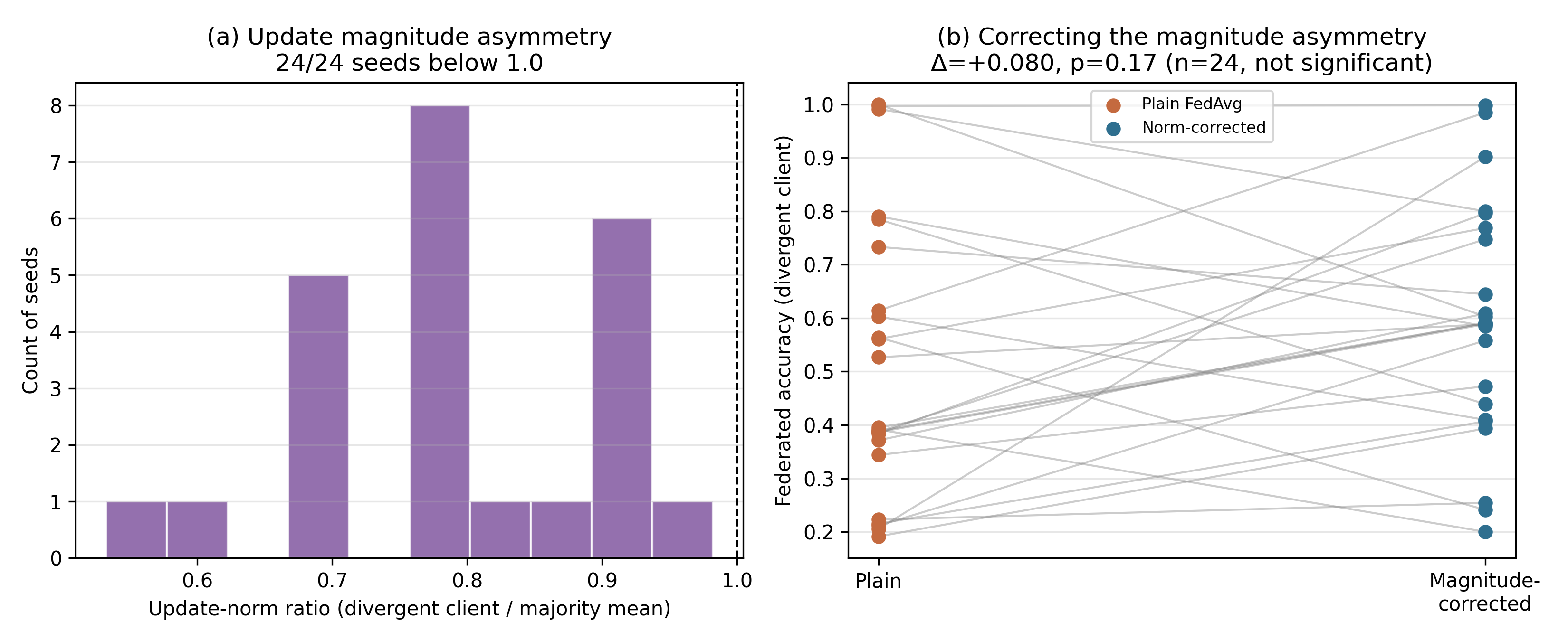}
\caption{(a) The structurally divergent client's shared-body update is smaller in magnitude than the federation majority's mean in 24 of 24 tested seeds, a highly consistent pattern under the tested GCN configuration. (b) Directly correcting this asymmetry helps on average but is not statistically significant at $n=24$, and does not reliably track harm magnitude.}
\label{fig:mechanism}
\end{figure*}

To test this, we held topology fixed at its most divergent, native setting across 24 independent seeds and measured update norms directly. The divergent client's update norm was smaller than the federation majority's mean update norm in \emph{every single seed} (ratio range 0.62--0.98). Under a null hypothesis of no true asymmetry, 24 out of 24 same-direction outcomes has probability $\approx6\times10^{-8}$. This is not a noisy or borderline finding under the configuration we tested (Figure~\ref{fig:mechanism}a). We want to be careful about how we phrase what we have actually shown here: it is a highly consistent empirical pattern in how GCN's degree-normalization interacts with structural divergence in this architecture and this training setup, not a claim we have shown to hold for GCN in general, across hidden widths, layer counts, or optimizers we did not test.

Having confirmed the asymmetry itself so decisively, the next question is whether it actually explains the accuracy harm it was proposed to explain. That is a distinct question from whether the asymmetry exists at all. Directly correcting it, by rescaling the divergent client's update to match the majority's mean norm each round with topology and every other factor held fixed, improved accuracy on average (+8.0 percentage points) but did not reach statistical significance even at this sample size ($p=0.17$ paired $t$-test, $p=0.13$ Wilcoxon signed-rank, $n=24$; Figure~\ref{fig:mechanism}b). More tellingly, the correlation between the measured asymmetry and the harm it was hypothesized to explain \emph{weakened} as sample size grew, from $r=-0.55$ at $n=8$ to $r=-0.18$ at $n=24$. This is the characteristic signature of a small-sample artifact regressing toward null as more data arrives, not of a real effect gaining statistical power.

So we are left holding two things at once, and we think both are worth reporting with equal weight rather than letting the more dramatic of the two overshadow the other. We have a mechanistic pattern in GCN aggregation under structural heterogeneity that held up in every single seed we tested, and no evidence that this pattern actually explains the accuracy harm that motivated looking for it in the first place. With causation still unresolved after two independent attempts to establish it, we turn in the next section to a different and more practical question. Even without a confirmed mechanism, can the association itself be turned into something that helps?

\section{Experiment 5: What We Tried to Fix the Problem, and What Happened}
\label{sec:exp4}

A diagnostic statistic, however well-validated, is only half useful without a way to act on it, and a plausible reader at this point in the paper would reasonably ask whether the association from Section~\ref{sec:exp1} can be turned into a working fix even absent a confirmed mechanism. This section reports, in the order we actually tried them, four attempts at doing exactly that. We present it as a sequence of attempts rather than as a single proposed method framed as if it had been the plan from the start, because the order matters to the story we are telling. Each attempt was motivated by what the previous one taught us, and the final attempt is the one that overturned what had looked, up to that point, like a genuine success.

\subsection{First attempt: our earlier carrier mechanism}
The most obvious first thing to try was a mechanism we had already built, before this study began, for a related purpose. This was the gradient-conflict regulation mechanism described in Section~\ref{sec:relwork}, which we refer to throughout as the carrier mechanism. It reweights and clips client updates by their cosine alignment with an evolving reference direction in raw parameter space. We tried it here as a first, off-the-shelf candidate, on the reasoning that if a general-purpose conflict-regulation mechanism already existed, it was worth checking before building anything new. It improved one of eight tested optimizer/architecture combinations significantly (GCN+FedProx: $71.3\%\to74.5\%$, $p=0.024$; Table~\ref{tab:realfed}), but was neutral everywhere else, and it made SAGE+SCAFFOLD significantly worse ($86.1\%\to81.9\%$, $p=0.013$). This is consistent with its cosine-similarity heuristic fighting the client-drift correction in SCAFFOLD rather than working alongside it. One win out of eight combinations tested was not the clean result we had hoped for, so we moved to something more targeted.

\subsection{Second attempt: calibrating FedProx directly from the diagnostic statistics}
If a generic, off-the-shelf mechanism was not the answer, the next logical step was to build something specifically informed by what we had actually validated, namely the degree and spectral divergence statistics from Section~\ref{sec:exp1}, rather than the carrier mechanism's generic, direction-consensus heuristic. We set each client's FedProx proximal strength $\mu_k$ and its federated aggregation weight as a function of its own measured divergence. Our first version \emph{relaxed} $\mu_k$ for divergent clients, on the reasoning that they were being over-constrained toward a global consensus that did not fit their own local optimum well. This made things worse rather than better. It underperformed plain fixed-$\mu$ FedProx overall (0.718 vs.\ 0.753 mean accuracy) and specifically hurt the very client it was meant to protect (sbm-dense-a: 0.565 vs.\ 0.816). Reasoning that we might simply have the sign wrong rather than the whole idea, we tried the opposite next: \emph{strengthening} $\mu_k$ for divergent clients instead. This recovered most of the lost ground (0.735--0.738, sbm-dense-a back up to 0.70--0.71), but still fell short of clearly beating the fixed baseline. Two attempts in, we had a sign flip that helped, but not yet a win worth reporting as a fix.

\subsection{Third attempt: a systematic search, evaluated on held-out seeds}
Rather than continue guessing at hyperparameters one adjustment at a time, we stepped back and ran a proper grid, evaluated in a way that would actually tell us whether we had found something real: $\mu_{\text{base}}\in\{0.01,0.02,0.04\}$ crossed with $\gamma\in\{0,1,2,4\}$, where $\gamma=0$ reproduces plain fixed-$\mu$ FedProx at that base value, giving us a like-for-like anchor inside the same grid rather than a separately-tuned comparison. We selected the best adaptive configuration and the best fixed anchor using seeds 0--2, then evaluated both on seeds 3--5, which had not been touched during selection, precisely so that any apparent win could not simply be an artifact of the selection process itself. The winning adaptive setting ($\mu_{\text{base}}=0.04$, $\gamma=4$) beat its matched anchor on the held-out seeds by $+3.5$ points, not significant at $n=3$ on its own, but $\gamma=4$ won at every one of the three tested $\mu_{\text{base}}$ values, which is a more systematic pattern than three independently lucky draws would typically produce. Encouraged by that consistency, we pushed $\gamma$ further, up to 16, and powered the final comparison to $n=10$ seeds; this turned into a clear, statistically significant result: 76.2\% (adaptive) vs.\ 71.8\% (fixed), $+4.3$ points, $p=0.001$ (Table~\ref{tab:mitigation}). At this point, the mechanism had passed a held-out-seed grid search and reached conventional statistical significance. By the standard playbook for reporting results of this kind, this is precisely where we would have stopped and called it a validated contribution.

\subsection{Fourth attempt: the control that changed the answer}
We did not stop there, because one detail kept nagging at us and seemed worth checking before we trusted the result. The winning configuration's average proximal strength across clients, 0.68, was far higher than the fixed baseline it had beaten, 0.02. That gap alone, entirely independent of any structural divergence-awareness, could plausibly explain the whole result. So we ran one further comparison, designed specifically to isolate that possibility. We tested a uniform proximal strength of 0.68, identical for every client and computed with no reference to structural divergence whatsoever. It used just the same average magnitude as the adaptive scheme, applied blindly to every client alike. This uniform control matched or slightly exceeded the divergence-calibrated version (76.9\% vs.\ 76.2\%, $p=0.57$, $n=10$; Table~\ref{tab:mitigation}). In other words, the entire statistically significant result from the previous attempt came from the original baseline simply being under-tuned at $\mu=0.02$, when something closer to $\mu\approx0.68$ was actually needed. It had nothing to do with structural divergence specifically.

\begin{table*}[t]
\caption{All four attempts, GCN+FedProx variants, real federation. $n=10$ for the final three rows (properly powered comparison); $n=3$ for earlier rows.}
\label{tab:mitigation}
\begin{tabular}{lrr}
\toprule
Method & Mean acc.\ & $\sigma_{\text{acc}}$ \\
\midrule
FedAvg & 0.723 & 0.190 \\
FedProx, fixed $\mu=0.01$ & 0.753 & 0.209 \\
SCAFFOLD & 0.705 & 0.223 \\
FedProx + carrier mechanism (attempt 1) & 0.730 & 0.197 \\
Divergence-calibrated, relaxed (attempt 2a) & 0.718 & 0.201 \\
Divergence-calibrated, strengthened (attempt 2b) & 0.735--0.738 & 0.19--0.21 \\
Grid-search fixed anchor, $\mu=0.02$ (attempt 3) & 0.718 & --- \\
Grid-search adaptive, $\mu_{\text{base}}{=}0.04,\gamma{=}16$ (attempt 3) & \textbf{0.762}$^*$ & --- \\
\textbf{Uniform control, $\mu=0.68$ (attempt 4)} & \textbf{0.769}$^\dagger$ & --- \\
\bottomrule
\end{tabular}
\\[2pt]
\footnotesize $^*$Significant vs.\ matched fixed anchor: $p=0.001$. $^\dagger$Matches or exceeds the adaptive mechanism: $p=0.57$ (not significant), showing no benefit from divergence-awareness beyond overall proximal strength.
\end{table*}

We think of this as the paper's second most important finding, after the association reported in Section~\ref{sec:exp1}, precisely because of how close it came to being reported as a success. A mechanism can pass every standard check in the federated learning evaluation playbook, including systematic grid search, held-out seeds, and conventional statistical significance, and still turn out to be nothing more than an under-tuned baseline once compared against a matched, structurally-blind control. We did not find this control applied in the comparable literature we reviewed, our own earlier work included, and we think it is worth adopting as standard practice going forward, independent of whatever this paper's other findings turn out to mean.

\section{Summary of Findings}
\label{sec:summary}

Having reported five experiments, several of which reversed conclusions drawn earlier in the same paper, we think a consolidated view is more useful to a reader than requiring reconstruction of the whole arc from five separate sections. Table~\ref{tab:summary} lists every claim tested across Sections~\ref{sec:exp1}--\ref{sec:exp4} against its outcome, so that the net result of this process, meaning what actually survived and what did not, is legible at a glance.

\begin{table*}[t]
\caption{Summary of every claim tested in this paper, its outcome, and status. ``Overturned'' denotes a result that was statistically significant under an initial test but did not survive a subsequent, more demanding control.}
\label{tab:summary}
\small
\begin{tabular}{p{2.6cm}p{3.6cm}p{1.5cm}}
\toprule
Claim tested & Outcome & Status \\
\midrule
Structural divergence is associated with negative transfer (real federation, $n=6$, initial) & $r=0.90$--$0.97$, FedProx & Supported \\
Degree divergence replicates at $n=20$ (threefold larger federation) & $r=0.49$--$0.53$, $p<0.03$; holds within synthetic-only subset & Supported, weaker \\
Spectral divergence replicates at $n=20$ & $r=-0.05$ to $0.15$, n.s.\ at full scale & Overturned (traced to reference-pool confound) \\
Density divergence is associated with negative transfer (synthetic, confounded design) & $r=0.50$, $p=0.034$, $n=18$ & Supported, later shown confounded \\
Homophily divergence is associated with negative transfer (density fixed) & $r=-0.28$ to $-0.37$, n.s. & Null \\
Feature-dimensionality mismatch is associated with negative transfer (topology fixed) & $r=0.54$, $p=0.011$, asymmetric & Supported (synthetic only) \\
Community granularity is associated with negative transfer & Confounded twice (capacity, then task-cardinality) & Excluded \\
Topology alone causally drives negative transfer (true intervention) & $r=0.07$--$0.08$, n.s., $n{=}800$ \& $n{=}3000$ & Null \\
GCN degree-normalization shrinks a divergent client's update magnitude & 24/24 seeds, $p\approx6\times10^{-8}$ & Consistent (mechanism only) \\
Correcting magnitude asymmetry rescues accuracy & $+8.0$ pts, $p=0.17$, $n=24$; correlation weakens with $n$ & Null \\
Carrier mechanism mitigates harm & 1/8 configs improved significantly, 1/8 degraded significantly & Fragile \\
Divergence-calibrated FedProx beats a tuned baseline & $+4.3$ pts, $p=0.001$ (initial) & Overturned ($p=0.57$ vs.\ uniform control) \\
\bottomrule
\end{tabular}
\end{table*}

\section{Discussion}
\label{sec:discussion}

Stepping back from the individual experiments, it is worth asking directly what the whole picture, taken together, actually establishes, and being honest that it is narrower than what a first pass at this problem, stopping after Section~\ref{sec:exp1} alone, would have suggested. We think that narrowness is itself the paper's most defensible contribution, precisely because it survived deliberate attempts to break it rather than being asserted and left untested. Structural negative transfer in federated GNN training is real and large; that part of the picture never changed across any of the five experiments. Degree divergence, specifically, is associated with it, cheaply and with no labels required, at a strength that held up, weaker but still statistically significant, when we expanded the test federation threefold and removed the contrast between real and synthetic domains entirely. Two things are easy to blur together here, and we want to keep them separate. The predictive availability of the statistic, meaning the fact that it exists and can be computed in advance before any training occurs, is well established by everything reported above. Its validated predictive generalization beyond the specific federations we studied is not, and would need a still larger and more diverse set of real federations, ideally spanning genuinely different graph domains rather than one real domain plus synthetic variation within it, to establish properly. In practical terms, this means a federation coordinator could compute degree divergence for a prospective client today and treat a high value as a warning sign worth investigating further, in the same spirit as an early screening signal in other fields, without yet treating it as a validated, generalizable predictor that has been shown to transfer to federations we have not tested.

The twenty-client expansion, beyond its direct bearing on the degree-versus-spectral question, carries a methodological lesson of its own that we think is worth drawing out explicitly, since it generalizes beyond this paper's specific statistics. Leave-one-out statistics of the kind we and others use to characterize a client's divergence from a federation are not fixed properties of the client considered alone. They are properties of the client relative to whatever pool happens to be measured against it at the time. When that pool's composition shifts, as it inevitably does whenever a federation grows or its membership changes over time, a statistic that looked like a strong, general signal can turn out, on closer inspection, to have been tracking the pool's composition as much as anything intrinsic to the client itself. We do not think this is unique to spectral divergence, or to this paper specifically; any diagnostic built on a leave-one-out or relative-to-the-rest comparison should be checked for whether it survives a change in the size and makeup of what it is being compared against, in the same way we checked ours here, somewhat by chance rather than by original design. We would not have found this without the client-pool expansion that a reviewer of an earlier draft pushed us to run, and we think the lesson generalizes as a caution well beyond our own specific statistics.

We now turn from what we found to what we still do not know. Despite a genuine and escalating attempt across three separate experiments, we could not establish a confirmed causal account for the surviving degree-divergence association, nor a mitigation that survives comparison against a properly-tuned, structurally-blind baseline. We see at least three non-exclusive explanations for the gap between a real association and absent causation, none yet distinguished by the data we have. The true causal driver may be feature-space or task separability rather than degree per se, incidentally correlated with degree in the federations we happened to study. The effect may be causal but require a combination of structural properties, or a scale of graph, beyond what our graph-surgery design was able to isolate. Or the association, though it survived our replication attempt, may still reflect a more diffuse notion of ``how atypical is this client'' that happens to correlate with degree without degree itself being the operative variable underneath it. The mechanistic pattern we did confirm in Section~\ref{sec:exp3b} is a plausible candidate for closing this gap eventually, but on the evidence we currently have, it does not close it.

The mitigation result from Section~\ref{sec:exp4} carries a methodological implication that extends well beyond this specific problem, and we think it is worth stating as a general caution rather than only as a result about our own mechanism. Divergence-calibrated and gradient-conflict mechanisms in federated graph learning are typically evaluated against a single fixed-hyperparameter baseline, without asking whether that baseline was tuned as carefully as the proposed method. Our results show this comparison can be systematically misleading. The mechanism we designed passed a held-out-seed grid search and reached conventional statistical significance against its baseline, and would, under standard reporting practice in this field, have been presented as a validated contribution without further question. Only a further control, which matched the mechanism's average hyperparameter magnitude in a structurally-blind uniform version, revealed that the entire effect was attributable to the baseline being under-tuned rather than to anything the mechanism itself contributed. We suspect this failure mode is not unique to our mechanism, and we recommend the same control as standard practice for the field going forward.

\section{Threats to Validity}

Having discussed what our results mean, we now turn to the specific ways they could be wrong or non-generalizable, organized by the standard categories of validity threat, so that a reader can weigh each concern against the specific claim it bears on.

\subsection{Construct Validity}
Negative transfer is operationalized as the accuracy delta between local-only and federated training under a matched compute budget; this is a reasonable choice but not the only possible one, and results could differ under a matched wall-clock or matched-convergence criterion instead, which we did not test. Structural divergence is operationalized via Wasserstein distance on degree sequences and adjacency-spectrum samples; these are two reasonable choices among several possible structural statistics, such as graphlet-based or curvature-based measures, and the strength of correlation we report is specific to this particular choice of statistic, not a claim that no other structural statistic could do as well or better.

\subsection{Internal Validity}
The graph-surgery design in Section~\ref{sec:exp3} is our strongest defense against confounding for the causal question, but it manipulates topology through a single generative path, interpolating stochastic block model edge probabilities, and we cannot rule out that a different topology-manipulation strategy would reveal a causal effect this particular design does not. The real-federation correlation in Section~\ref{sec:exp1} cannot rule out unmeasured confounds tied to client identity, such as domain, feature dimensionality, or dataset curation practices, despite its leave-one-out construction, which is precisely why Sections~\ref{sec:exp2}--\ref{sec:exp3} attempt, and only partially succeed at, isolating structure as an independent variable. The mitigation search in Section~\ref{sec:exp4}, while including a held-out-seed grid and the uniform-baseline control that proved decisive, explored a bounded hyperparameter space and a single mechanism family, proximal-term calibration; we cannot rule out that a differently-structured mechanism, informed by the same validated predictors, would survive the same control where ours did not.

\subsection{External Validity}
The real-federation correlation was tested at two scales, six and twenty clients, but within one family of federations, consisting of three fixed real citation graphs plus a growing set of synthetic structural proxies, rather than across genuinely independent federations built from different real graph domains. The twenty-client result is a real strengthening of the evidence, but it is not the same as replicating the finding on an entirely separate federation, and the domain-confound diagnosis in Section~\ref{sec:exp1b} depends on synthetic clients sharing a generative process, which may not hold for a federation built from independently-sourced real graphs. All experiments use two architectures, GCN and GraphSAGE, one task type, node classification, and graph sizes between 500 and 20{,}000 nodes. Generalization to larger federations spanning tens to hundreds of clients, to other GNN variants such as attention-based or spectral methods, to other task types such as link prediction or graph classification, and to graph domains beyond citation, co-purchase, and co-authorship proxies remains untested and should not be assumed.

\subsection{Statistical Conclusion Validity}
Several individual tests in this paper are run at small $n$. These include the original real-federation correlations at $n=6$, later expanded to $n=20$ in Section~\ref{sec:exp1b}, and several early mitigation comparisons at $n=3$ before being deliberately powered up to $n=10$--$24$. We report exact $p$-values and effect sizes throughout rather than significance thresholds alone, and, notably, three of our own mid-study conclusions were reversed after adding more data. These were the real-federation correlation itself once expanded to twenty clients, the magnitude-asymmetry correlation in Section~\ref{sec:exp3b}, and the divergence-calibrated mitigation result in Section~\ref{sec:exp4}. We take this recurring pattern as a caution about this paper's own findings specifically, not only about the field's more generally. We did not apply a formal multiple-comparisons correction across the more than twenty statistical tests reported across Sections~\ref{sec:exp1}--\ref{sec:exp4}; some fraction of the nominally significant results, including the surviving degree-divergence association itself, may not hold up under such a correction or under further replication, and we flag this explicitly rather than treating any single $p$-value as independently conclusive on its own.

\section{Reproducibility}

All code used to construct the federation, compute the structural divergence metrics, run every experiment in Sections~\ref{sec:exp1}--\ref{sec:exp4} (including the twenty-client expansion in Section~\ref{sec:exp1b}), and generate every figure and table in this paper will be made publicly available in a code repository upon publication, organized by experiment, with a top-level README describing how to reproduce each result and the exact random seeds used for every reported number. The real citation graphs (Cora, CiteSeer, PubMed) are the standard, publicly-available Planetoid splits~\cite{yang2016}; the synthetic structural-proxy graphs, including the fourteen additional clients used in Section~\ref{sec:exp1b}, are generated from documented stochastic block model parameters with fixed seeds, and the generation code itself, not only the resulting data, will be included, so that the controlled sweeps in Sections~\ref{sec:exp2}--\ref{sec:exp3} and the client-pool expansion in Section~\ref{sec:exp1b} can be regenerated exactly or extended to new parameter ranges and new pool sizes.

\section{Future Work}

Four directions follow naturally from where this paper's evidence currently stops. The most direct is disentangling the candidate explanations raised in Section~\ref{sec:exp3b} for why a real association does not translate into confirmed causation. In particular, this means isolating feature-space or task separability as an independent causal variable using the same graph-surgery methodology applied here to topology. A second direction is testing whether the uniform-baseline control we introduce in Section~\ref{sec:exp4} overturns other divergence-aware or gradient-conflict mechanisms already published in the federated graph learning literature, when applied to their original evaluation settings rather than only to our own. A third, motivated directly by the limitation named in Section~\ref{sec:exp1b}, is replicating the degree-divergence association on federations built from genuinely independent real graph domains. This means separately-sourced real graphs from, for instance, healthcare, financial, or web-scale settings, rather than the synthetic expansion within one family that we were able to do here. Such a replication would both test generalization properly and check whether the reference-pool-composition confound we identified for spectral divergence has an analogue we have not yet detected for degree divergence. A fourth, suggested by the confound itself, is testing whether other leave-one-out or relative-to-the-rest diagnostics already in use elsewhere in the federated learning literature are similarly sensitive to reference-pool composition, which would make this paper's methodological lesson considerably more general than a single case study.

\section{Conclusion}

We set out to test whether federated graph neural network training silently harms structurally atypical participants, and whether that harm could be predicted, explained, and fixed. What we can report, honestly, is a real, moderate, and partially-replicated association between degree divergence and structural negative transfer. It survived an expansion of the test federation and the removal of a specific confound, after an initially much stronger correlation on a smaller sample did not survive intact. We report this alongside equally rigorous negative results. These comprise a multi-angle failure to confirm a causal mechanism across three independent attempts, and a mitigation search that appeared successful until a necessary control revealed that it was not. We believe all of this belongs in the published record together, not just the parts that worked. The surviving diagnostic signal is usable today as an early warning sign, though not yet as a validated predictor, independent of whether its mechanism is ever confirmed; the honestly-reported negative and partially-negative results narrow the search space for whoever extends this work next, rather than leaving them to rediscover the same dead ends. Finally, the two confounds we caught in our own results, a small-sample domain confound in Section~\ref{sec:exp1b} and a baseline-tuning confound in Section~\ref{sec:exp4}, are cautions that we think extend well beyond this specific paper, to how federated heterogeneity research is evaluated more generally.

\end{document}